\pdfoutput=1

\documentclass[11pt]{article}

\usepackage[final]{acl}

\usepackage{times}
\usepackage{latexsym}

\usepackage[T1]{fontenc}

\usepackage[utf8]{inputenc}

\usepackage{microtype}

\usepackage{inconsolata}

\usepackage{graphicx}

\usepackage{hyperref}
\usepackage{url}
\usepackage{booktabs}       
\usepackage{amsfonts}       
\usepackage{nicefrac}       
\usepackage{microtype}      
\usepackage{xcolor}         
\usepackage{multicol,multirow}
\usepackage{enumitem}
\usepackage{subcaption}
\usepackage{amsmath}
\usepackage{makecell}

\title{LLM4DistReconfig: A Fine-tuned Large Language Model for Power Distribution Network Reconfiguration}

\author{
  Panayiotis Christou$^{1}$, 
  Md. Zahidul Islam$^{1,*}$, 
  Yuzhang Lin$^{1}$, 
  Jingwei Xiong$^{2}$ \\
  $^1$Tandon School of Engineering, New York University, Brooklyn, NY \\
  $^2$University of California, Davis, CA \\
  \texttt{\{pc2442, mi2502, yl12252\}@nyu.edu}, \texttt{jwxxiong@ucdavis.edu} \\
  $^*$Corresponding author: \texttt{mi2502@nyu.edu}
}

\begin{document}
\maketitle
\begin{abstract}


Power distribution networks are evolving due to the integration of distributed energy resources (DERs) and increased customer participation. To maintain optimal operation, minimize losses, and meet varying load demands, frequent network reconfiguration is necessary. Traditionally, the reconfiguration task relies on optimization software and expert operators, but as systems grow more complex, faster and more adaptive solutions are required without expert intervention. Data-driven reconfiguration is gaining traction for its accuracy, speed, and robustness against incomplete network data. Large language models (LLMs), with their ability to capture complex patterns, offer a promising approach for efficient and responsive network reconfiguration in evolving complex power networks.

In this work, we introduce \textbf{LLM4DistReconfig}, a deep learning-based approach utilizing a fine-tuned LLM to solve the distribution network reconfiguration problem. By carefully crafting prompts and designing a custom loss function, we train the LLM with inputs representing network parameters such as buses, available lines, open lines, node voltages, and system loss. The model then predicts optimal reconfigurations by outputting updated network configurations that minimize system loss while meeting operational constraints. Our approach significantly reduces inference time compared to classical algorithms, allowing for near real-time optimal reconfiguration after training. Experimental results show that our method generates optimal configurations minimizing system loss for five individual and a combined test dataset. It also produces minimal invalid edges, no cycles, or subgraphs across all datasets, fulfilling domain-specific needs. Additionally, the generated responses contain less than 5\% improper outputs on seen networks and satisfactory results on unseen networks, demonstrating its effectiveness and reliability for the reconfiguration task.
\end{abstract}

\section{Introduction}
\label{sec:introduction}

Power distribution network reconfiguration is crucial for maintaining operational efficiency, reliability, and adaptability in modern power networks. The integration of distributed energy resources (DERs), such as renewable energy sources like solar and wind, introduces variability and uncertainty, necessitating sophisticated control strategies to balance loads, minimize power losses, and maintain voltage stability \cite{sultana2016review}. Effective network reconfiguration enhances the network's ability to accommodate these renewable sources, optimize energy flow, and ensure resilience against disturbances, supporting the transition toward a more sustainable and flexible power infrastructure.

Traditional optimization methods for network reconfiguration, while effective, often suffer from computational complexity and scalability issues, especially as power networks grow in size and complexity \cite{mishra2017comprehensive}. Large Language Models (LLMs) have emerged as a promising alternative, capable of learning complex patterns and dependencies within large datasets. By leveraging their ability to handle diverse inputs and generate structured outputs, LLMs can provide rapid, data-driven solutions for network reconfiguration, reducing the need for iterative computations typical of conventional approaches.

LLMs, such as GPT \cite{brown2020languagemodelsfewshotlearners} and BERT \cite{devlin2019bertpretrainingdeepbidirectional}, have revolutionized natural language processing and have shown remarkable success in tasks involving understanding, generation, and reasoning over textual data. Their applications have extended beyond text-related tasks to complex decision-making domains, leveraging their capacity to process vast amounts of data and generate coherent, context-aware predictions \cite{minaee2024largelanguagemodelssurvey}. In power systems, the potential of LLMs is explored in predictive maintenance, optimal power-flow, and decision-making processes that involve large-scale, multi-objective considerations \cite{huang_large_2023, choi2024egridgpt}.

However, applying LLMs to the power distribution network reconfiguration task presents unique challenges. The critical nature of this task means that errors can lead to system instability or inefficiencies. It involves multiple sub-tasks, such as assessing system voltages, managing losses, and adhering to physical laws, all of which require precise decision-making \cite{sultana2016review}. Fine-tuning LLMs specifically for reconfiguration tasks is essential to ensure that the models can effectively handle these complexities and contribute meaningful solutions.

Existing literature on network reconfiguration has focused on various optimization techniques. Early work by \cite{merlin_back_1974} introduced a heuristic search method for loss minimization, formulating the problem as finding a minimal-loss spanning tree. \cite{civanlar_1988} proposed a simplified branch exchange method to reduce losses in distribution feeders. Heuristic methods were further advanced by \cite{gomes_2005}, who developed algorithms capable of handling larger and more complex systems. Evolutionary approaches, such as genetic algorithms applied by \cite{roux_2012}, have also been explored for load balancing and multi-objective optimization.

Recent studies have begun integrating machine learning techniques with optimization algorithms \cite{ji2021real}. For instance, \cite{li2021many} presented a deep reinforcement learning framework for multi-objective network reconfiguration under varying network conditions. Despite these advancements, the application of LLMs to reconfiguration task remains unexplored.


\paragraph{Our Contributions:}

In this paper, we introduce \textbf{LLM4DistReconfig}, a novel approach utilizing fine-tuned large language models to solve the power distribution network reconfiguration problem. To the best of our knowledge, this is the first work to fine-tune LLMs with a custom loss function for this task. Our key contributions are as follows:

\begin{itemize}[leftmargin=0.5cm]
    \item We fine-tune LLaMA 2 \cite{touvron2023llama2openfoundation} and LLaMA 3 \cite{dubey2024llama3herdmodels} models to optimize network reconfiguration in response to changing system demands, utilizing the expanded context window in LLaMA 3.1 (120k tokens) to handle larger network sizes. Custom datasets are created for various network sizes in ChatML format, optimized for power networks by reducing precision where necessary and removing irrelevant details.

    \item Our approach refines prompt instructions to prevent issues such as invalid line selections, cycles, and subgraphs, which violate power domain-specific constraints. We ensure the outputs adhere to a structured format (e.g., system loss, line tuples) and guide the model to retain previous configurations when new ones do not reduce system losses.

    \item We design post-processing pipelines for both training and inference to parse model outputs and calculate a custom loss based on the presence of invalid lines, cycles, and subgraphs in the reconfiguration task. A penalty-based system in the custom loss function helps the model learn to avoid these issues, improving line selection accuracy and enforcing domain-specific constraints.

    \item To ensure robustness, the final outputs undergo post-processing to eliminate responses that could lead to system failures. The model also supports interactive modifications through chat, allowing users to refine or guide responses to meet custom output requirements.
\end{itemize}

Our work demonstrates the potential of LLMs in solving complex optimization problems in power networks, highlighting the importance of combining prompt engineering with customized training objectives when adapting LLMs to specialized domains. 
The dataset and codebase used in this paper have been made publicly available on GitHub\footnote{\url{https://github.com/panaschristou/LLM4DistReconfig}} to facilitate the replication of our results and encourage further contributions.

\section{Reconfiguration Problem and Large Language Models}
\label{sec:overview}

\subsection{Reconfiguration Problem}

The power distribution network reconfiguration problem involves finding an optimal topology of the distribution network to minimize system losses and improve operational performance while satisfying all operational constraints. The distribution network is modeled as a graph $G(N, E)$, where nodes $N$ represent buses (load points, substations, etc.) and edges $E$ represent distribution lines or feeders. Certain edges, denoted as $E_{\text{sw}} \subseteq E$, are equipped with switches that can be opened or closed to alter the network configuration.

The objective is to determine the status (open or closed) of these switchable lines to minimize power losses, subject to several constraints. These include balancing the total power generated and demanded, ensuring power flows do not exceed line capacities, maintaining a radial and connected network (i.e., no cycles and subgraphs), and keeping voltage levels within acceptable bounds. The detailed formulations are provided in \ref{appendix:formulation}.

Mathematically, the reconfiguration problem is a complex combinatorial optimization problem and is classified as NP-hard \cite{li2021many}. The vast number of possible configurations, especially in large-scale networks, makes exhaustive search infeasible. Therefore, efficient and effective solution methods are essential for practical applications.

\subsection{Fine-Tuning LLMs for the Reconfiguration Task}

Applying LLMs to the network reconfiguration problem offers a novel approach to address its inherent complexities. However, directly applying pre-trained LLMs to this domain is insufficient due to the highly specialized and technical nature of the task. Fine-tuning is necessary to adapt the LLM to understand the specific constraints and operational principles of power distribution networks.

Fine-tuning the LLM involves training it on domain-specific datasets that include examples of network configurations, operational scenarios, and associated outcomes. Through this process, the LLM learns to understand network topology, incorporate operational constraints, optimize objectives, and generate feasible configurations.

By fine-tuning the LLM for the reconfiguration task, we can harness its ability to generalize from data, enabling it to propose effective reconfiguration strategies without exhaustive computation or extensive expert intervention. This approach can facilitate near real-time decision-making and make the reconfiguration process more feasible.

\subsection{Dataset Preparation}

A critical component of fine-tuning LLMs for domain-specific tasks is the availability of a high-quality dataset. In the context of network reconfiguration, real-world operational data may be scarce or confidential. To overcome this limitation, we generate a synthetic dataset that simulates realistic operational scenarios.

We utilize established power system simulation tools, such as MATPOWER \cite{zimmerman2010matpower}, to model and simulate various network configurations and loading conditions. Specifically, we consider standard IEEE distribution test feeders, including the 33-node, 37-node, 69-node, 84-node, and 136-node networks \cite{harsh2023simple}. These test cases provide diverse network topologies and complexities for comprehensive training.

For each network, we simulate various loading scenarios by varying the demand at different buses using open-source load data \cite{noauthor_smartmeter_nodate}. The simulation generates corresponding optimal reconfiguration solutions using the optimization algorithm documented in the literature \citet{mishra2017comprehensive}. Each data sample includes network topology, operating voltages and system loss,  operating configuration (i.e., open lines), the resulting configuration, power losses, voltage profiles.

\begin{figure*}[t]
    \centering
    \includegraphics[width=\linewidth]{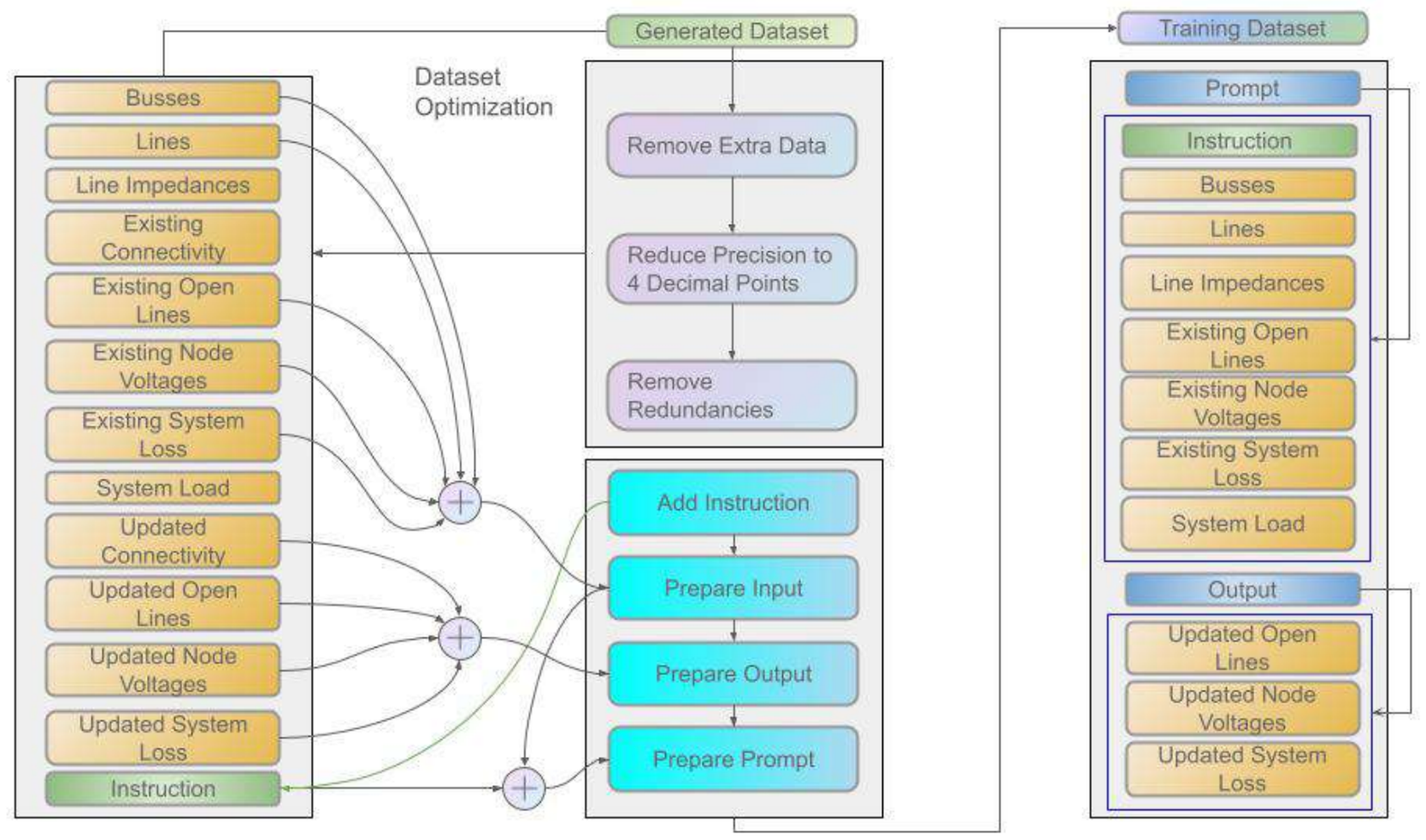}  
    \caption{Process of Dataset Generation for Fine-Tuning the LLM}
    \label{fig:dataset_creation}
\end{figure*}

The dataset is formatted to be compatible with the LLM's input requirements, ensuring that the model can effectively learn from the data. The complete dataset processing pipeline is shown in Figure \ref{fig:dataset_creation}. As shown in Figure \ref{fig:dataset_creation} we remove redundancy, and thus reducing the number of tokens in the prompt, by removing the existing and updated connectivity matrices since their information is already conveyed by the busses and lines. We also reduce the precision as outlined in section \ref{sec:framework}. We also represent the nodes (buses) as a single inferred number (Nth node), starting from 1, rather than a list of numbers for a more simplified node representation. We append instructions to each sample as part of the prompt as detailed in Appendix \ref{appendix_sample_responses}. By generating a large number of such samples across different networks and scenarios, we provide the LLM with sufficient information to capture the complex relationships inherent in the reconfiguration problem.

To prepare for training, the processed dataset is converted to ChatML format for training. The input is comprised with the following columns:

\begin{itemize}[leftmargin=0.5cm]
    \item \textbf{Buses:} The nodes of the power grid.
    \item \textbf{Lines:} The edges of the power grid.
    \item \textbf{Line Impedances:} Edge features representing the characteristics of the lines.
    \item \textbf{Existing Open Lines:} The initial set of deactivated lines.
    \item \textbf{Existing Node Voltages:} The initial voltages levels at each node.
    \item \textbf{Existing System Loss:} The total system loss in the initial configuration.
    \item \textbf{System Load:} The load on each node at a specific time.
\end{itemize}

\textbf{Example:} Buses: 33, Lines: [(1,2), (2,3), ...], Line Impedances: [0.00064569, 0.00, ...]

The output is similarly structured and includes the reconfigured values. The output is comprised with the following columns:

\begin{itemize}[leftmargin=0.5cm]
    \item \textbf{Updated Open Lines:} The new set of deactivated lines.
    \item \textbf{Updated Node Voltages:} The updated voltage levels at each node.
    \item \textbf{Updated System Loss:} The total system loss after reconfiguration.
\end{itemize}

The input and instruction are concatenated to form the prompt. An example task description including prompt and input is provided in Table \ref{table:task_description} in Appendix \ref{appendix_sample_responses}. The output remains separately structured to allow for accurate training. More dataset details can be found in Appendix \ref{appendix_dataset_details}.





\section{Incorporating Domain-Specific Constraints into LLMs}
\label{sec:finetuning}

\subsection{Instruction Prompt Refinement}

Adapting LLMs to domain-specific tasks requires careful guidance to ensure adherence to specialized constraints. In the context of power distribution networks, characterized by attributes such as the number of buses, lines, operating voltages, and system losses, LLMs must generate outputs that are not only syntactically correct but also operationally feasible.

To achieve this, instruction prompts were iteratively refined to embed domain knowledge and constraints directly into the model's input. The initial prompts simply described the reconfiguration problem, representing the network as a graph with buses as nodes and lines as edges. However, the model often generated invalid configurations, including nonexistent lines and cyclic graphs, indicating a gap in understanding the domain's constraints.

Motivated by these observations, additional instructions were incorporated:

\begin{itemize}[leftmargin=0.5cm]
    \item \textbf{Constraint on Valid Lines:} Only lines provided in the input data should be considered to prevent the generation of invalid or nonexistent lines. This helps the model focus on feasible reconfigurations within the actual network topology.
    \item \textbf{Ensuring Radial Connectivity:} The output graph must be a single connected component without cycles, reflecting the radial nature of power distribution networks. This guides the model toward generating operationally viable configurations.
    \item \textbf{Acyclic Graph Specification:} The number of closed lines must equal the number of nodes minus one, which mathematically enforces an acyclic (tree) structure. This instruction addresses the issue of residual cycles in the model's output.
    \item \textbf{Operational Constraints and Output Formatting:} Practical considerations include instructing the model not to reconfigure the network if the inferred system loss increases and providing explicit output format requirements. This ensures that the model's outputs are not only correct but also practically useful.
\end{itemize}
Each refinement was motivated by specific shortcomings in the model's performance, aiming to incrementally improve its adherence to domain constraints and operational requirements. Figure \ref{fig:dataset_creation} shows how the initial dataset goes through the steps of the processing by reducing precision, removing columns, adding the task, generating the prompt and eventually getting to the final dataset used for training and evaluation.

\subsection{Custom Loss Function}

Even with refined prompts, the model sometimes produced outputs violating domain constraints. To further enforce these constraints during training, a custom loss function was developed, comprising three key components:

\paragraph{Cycle Loss}

Cycles in the output graph $G_{\text{output}} = (V, E_{\text{available}} \setminus E_{\text{output}})$ obtained from model responses are undesirable, as they violate the radial topology of power distribution networks. The cycle loss penalizes the presence of such cycles:
\[
\mathcal{L}_{\text{cycle}} = |C(G_{\text{output}})|
\]
where $C(G_{\text{output}})$ is the set of cycles in $G_{\text{output}}$. This component encourages the model to generate acyclic graphs.

\paragraph{Subgraph Loss}

Disconnected subgraphs indicate a network that is not fully connected, which is operationally infeasible. Subgraph loss penalizes the existence of multiple connected components:
\[
\mathcal{L}_{\text{subgraph}} = k - 1
\]
where $k$ is the number of connected components in $G_{\text{output}}$. This loss component drives the model toward producing a single, fully connected network.

\paragraph{Suboptimal Configuration Loss}

The inclusion of lines in generated open lines not present in the optimal open lines can lead to suboptimal configurations. The suboptimal configuration loss penalizes such occurrences  by subtracting generated open lines ($E_{output}$) from optimal open lines ($E_{optimal}$):
\[
\mathcal{L}_{\text{subconfig}} = |E_{\text{output}} \setminus E_{\text{optimal}}|
\]

\paragraph{Total Loss}

The total loss combines the standard training loss $\mathcal{L}_{\text{reg}}$ with custom loss components, each scaled by a factor $\lambda_i$ to balance their impact:

\[
\mathcal{L}_{\text{total}} = \mathcal{L}_{\text{reg}} + \lambda_1 \mathcal{L}_{\text{cycle}} + \lambda_2 \mathcal{L}_{\text{subgraph}} + \lambda_3 \mathcal{L}_{\text{subconfig}}
\]
This comprehensive loss function guides the model to produce outputs that are not only syntactically correct but also conform to domain-specific operational constraints.

\subsection{Fine-Tuning Process and Loss Scaling}

The fine-tuning process involves presenting the refined prompts to the model and analyzing its outputs to compute the custom loss components. Model output is obtained by parsing the output, detailed in \ref{appendix:challenges}. To ensure stable training and effective learning, scaling of the loss components is essential:

\begin{itemize}[leftmargin=0.5cm]
    \item \textbf{Cycle Loss Scaling:} The cycle loss is divided by the number of available lines to prevent it from dominating the total loss, especially in cases where the number of potential cycles is large.
    \item \textbf{Subgraph and Suboptimal Configuration Loss Scaling:} Both losses are divided by the total number of predicted lines, normalizing their impact relative to the size of the output and ensuring they contribute appropriately to the total loss.
\end{itemize}
In instances where the model outputs are incorrectly formatted or fail to meet the specified constraints, termed as \textbf{improper responses}, maximum penalties are applied by setting all custom loss components to 1.
 This strategy encourages the model to adjust its outputs in future iterations to align with the domain requirements.

\begin{figure}[t]
    \centering
    \includegraphics[width=\linewidth]{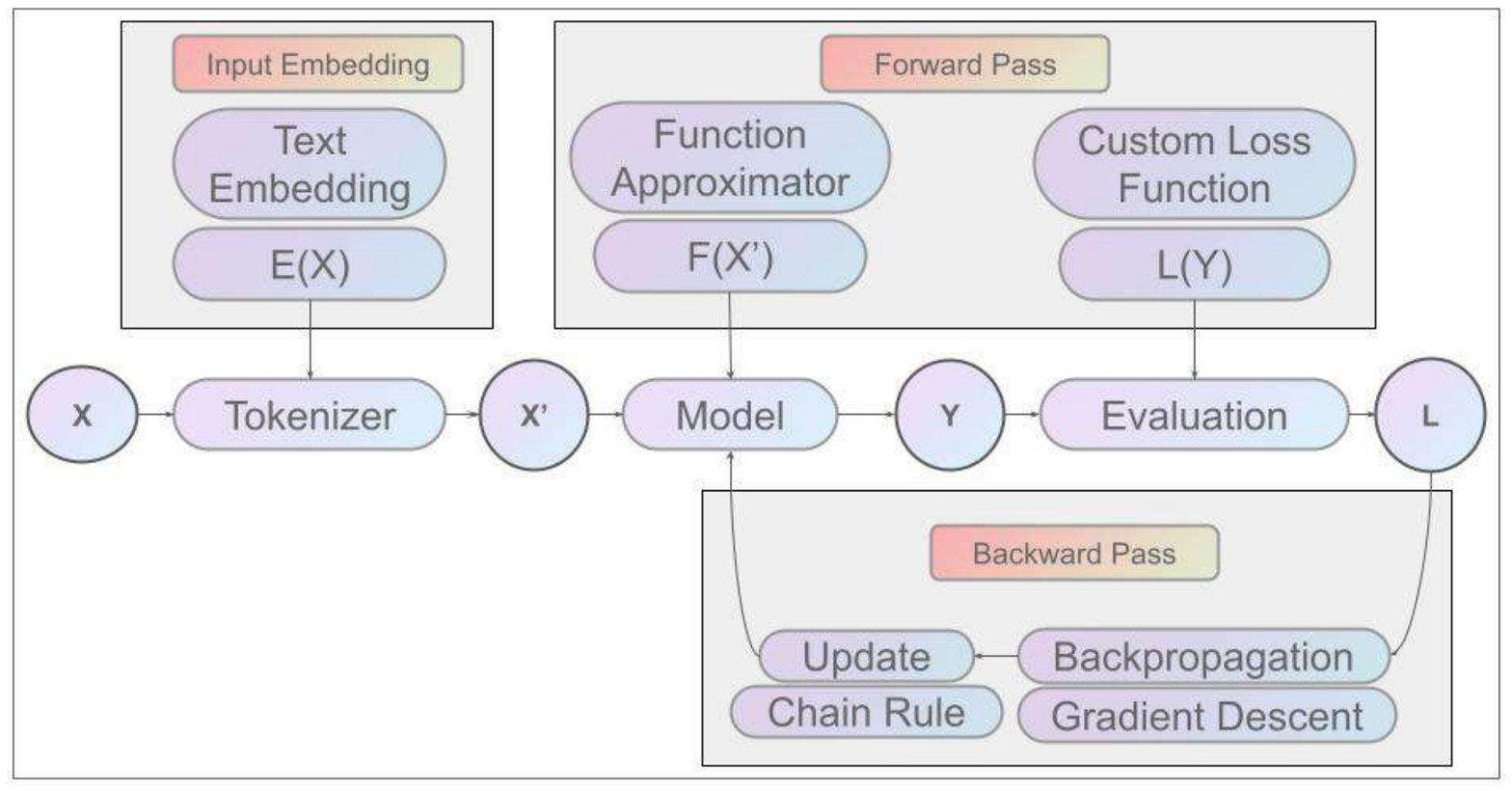}  
    \caption{Full Training Pipeline Diagram}
    \label{diagram:full_pipeline_diagram}
\end{figure}

For effective training, a learning rate of 0.0002 was employed during the initial stages. A cosine learning rate scheduler was utilized to adaptively adjust the learning rate throughout training, promoting convergence and enhancing the model's performance. Figure \ref{diagram:full_pipeline_diagram} shows the full end to end training pipeline of the model, going from the input to the embedding, to the model, then evaluating the output and calculating the gradients to update the model, effectively performing gradient descent through forward and backward passes.

By integrating domain-specific constraints directly into the instruction prompts and loss function, the model is guided to generate outputs that are both accurate and operationally viable within the context of power distribution networks. 



\section{Experimental Setup}
\label{sec:framework}


\subsection{Dataset Generation and Optimization}
We generate our dataset using IEEE Y Bus Systems with sizes ranging from 33 to 136 buses (denoted as 33N, 37N, 69N, 84N, and 136N). For each network, the dataset is represented as $X^{m \times n}$, where $n=17,520$ is the number of samples and $m=12$ is the initial number of features. To optimize the dataset, we reduce the feature set by dropping redundant connectivity data, resulting in a reduced dataset with $m=10$ features:
$$ X^{m \times n} = X^{17520 \times 12} \rightarrow X^{17520 \times 10} $$

Additionally, we scale the number of buses from a list of comma-separated values to a scalar, representing the total number of buses. Precision is reduced to five decimal places for all variables:
$$ v' = \text{round}(v, 5) $$

\subsection{LLM Training and Inference}
Initially, we experimented with GPT-3, but its small context window (2049 tokens) and proprietary nature led us to switch to LLaMA-2 (7B parameters) with a 4096-token context window. However, even this model posed limitations due to its context size. LLaMA-3.1, with a 120,000-token context window and optimized training features such as Flash Attention 2, solved these issues, as explained in \ref{appendix:challenges}.

\begin{figure}[t]
    \centering
    \includegraphics[width=\linewidth]{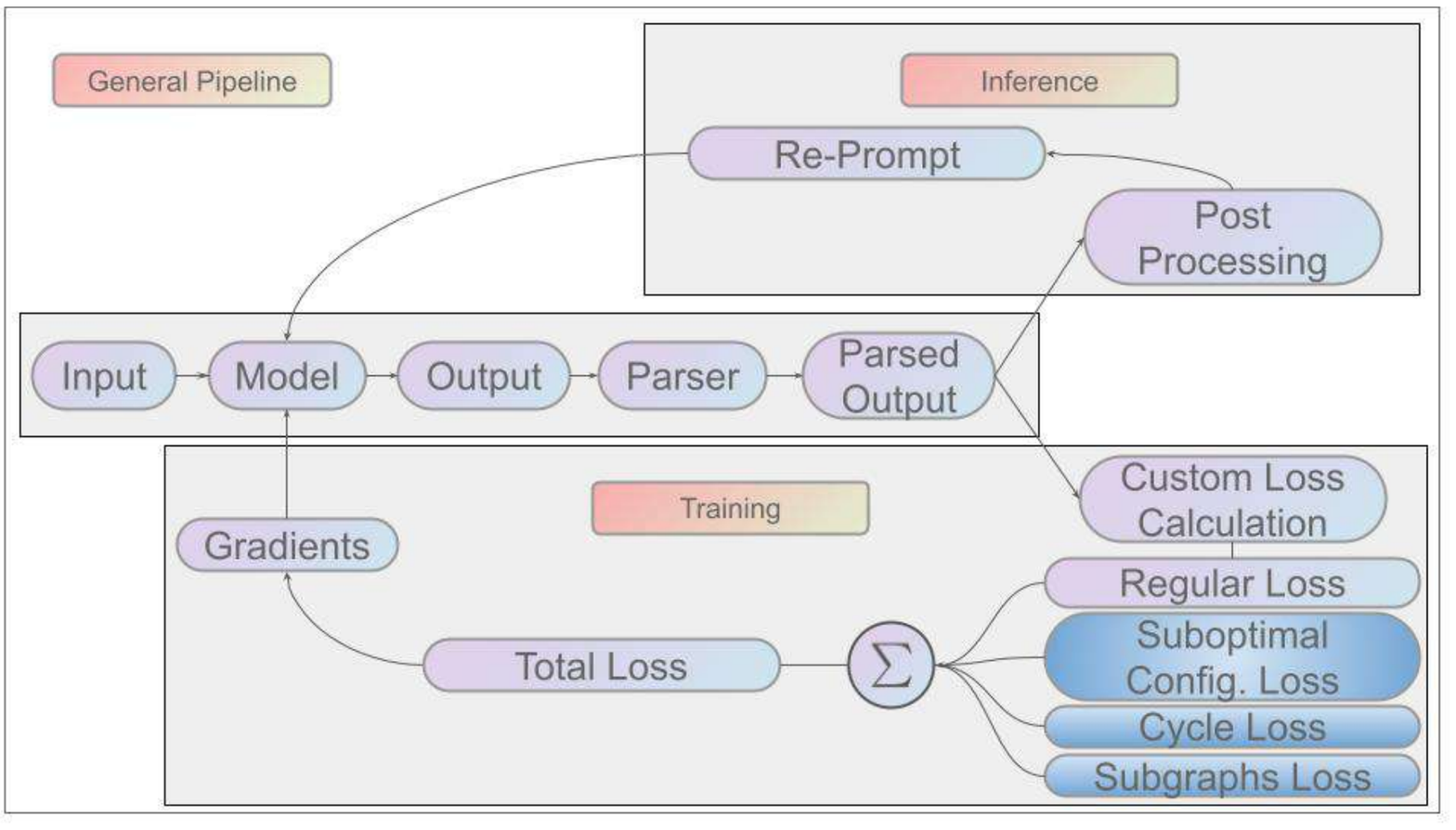}  
    \caption{Training \& Inference Diagram}
    \label{diagram:training_inference_diagram}
\end{figure}

We fine-tuned the LLaMA-3.1 model, our flagship fine-tuned model, on a combined dataset (33N, 69N, and 84N) with 52,560 samples, split equally into training, validation, and test sets. The model was trained for 30 epochs on a V100 GPU (20GB VRAM) over 285 hours using the LLaMA-3.1 tokenizer and AutoTrain from Huggingface. While we considered using A100 GPUs (80GB VRAM) for larger batch sizes and distributed training, resource constraints led us to prioritize multiple jobs across different datasets. The model was trained in increments of 10 epochs, allowing us to evaluate model performance after each training stage. The full training and inference pipeline is shown in diagram \ref{diagram:training_inference_diagram}. We can see the main backbone of the pipeline stays the same and only the post generation part changes accordingly for custom loss calculation during training or post-processing re-prompting during inference. A full list of hyperparameters is provided in \ref{appendix_hyperparameters}. 

\subsection{Model Testing and Generalization}
We evaluated the model's performance on the test sets for 33N, 69N, and 84N networks, as well as unseen 37N and 136N networks to assess generalization. The metrics included the number of suboptimal configuration, cycles, subgraphs, and average inference time. The average inference time per network size and per maximum number of new tokens generated is provided in Appendix \ref{appendix:inference_times}. Testing included both in-distribution generalization (networks within the training size range) and out-of-distribution generalization (networks outside the training size range).

This experimental setup enabled us to systematically evaluate the model’s performance on various network sizes and configurations while balancing computational resources and accuracy.

\section{Case Studies}
\label{sec:case_studies}

\subsection{Effectiveness of augmented prompts and custom loss}
\begin{figure}
    \centering
    \includegraphics[width=0.8\linewidth]{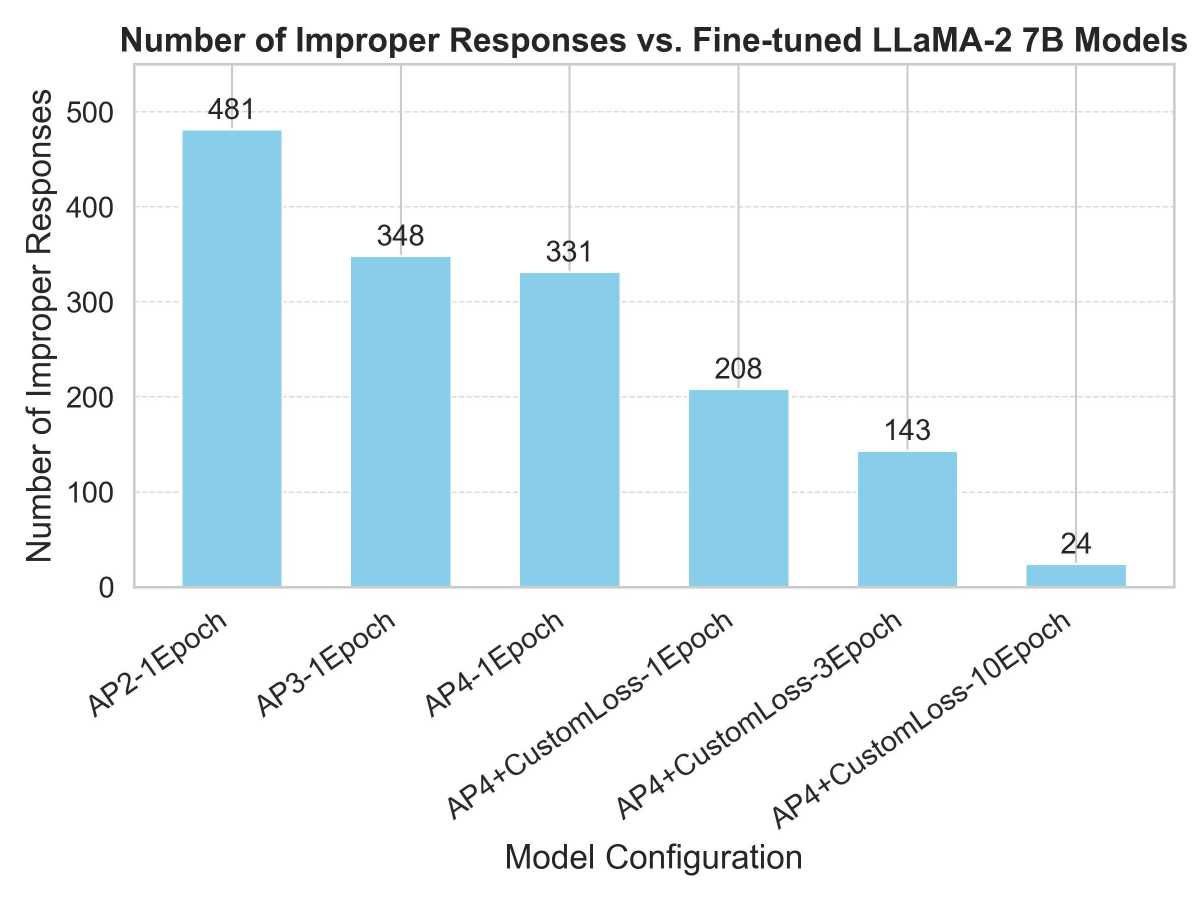}
    \caption{Comparison of fine-tuned LLaMA-2 models on generating improper responses, illustrating the impact of the custom loss function, augmented prompts, and training epochs on model performance.}
    \label{fig:llama2-customloss}
\end{figure}
Below, we present the performance of different models, highlighting the effects of prompt augmentation, the addition of custom loss components, and the impact of training for more epochs. 
While the cycle and subgraph issues were addressed with augmented prompts, the number of improper responses remained high.

In Figure \ref{fig:llama2-customloss}, we show the performance of the fine-tuned LLaMA-2 7B model with different iterations of augmented prompts (APx), custom loss, and varying training epochs. The figure demonstrates that the model generated fewer improper responses as more instructions were added. Specifically, when moving from iteration 2 to 4, the improper responses were reduced from 481 to 331 out of 500 test samples on the combined dataset, representing approximately a 31\% improvement. After adding custom loss alongside the augmented prompts and training the model, the improper responses further decreased to 208. This number continued to decline as we increased the training epochs from 1 to 3 to 10. Finally, with 10 epochs of training, augmented prompts, and custom loss, the number of improper responses dropped to 24 out of 500 test samples—less than 5\%. The figure illustrates the effectiveness of both prompt augmentation and the custom loss function. We observed a similar trend for our flagship fine-tuned LLaMA-3.1 8B model.

\begin{figure}
    \centering
    \includegraphics[width=0.8\linewidth]{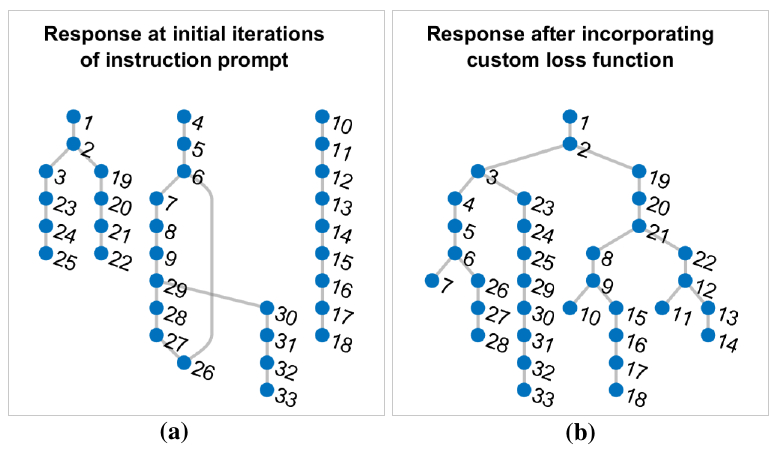}
    \caption{Network configuration: (a) without and (b) with the custom loss function.}
    \label{fig:33N_compare}
\end{figure}

Figure \ref{fig:33N_compare} shows a sample LLM-generated network configuration. In the initial iterations of instruction prompts, cycles and subgraphs were observed in the generated configuration, as seen in Figure \ref{fig:33N_compare}(a). As we trained the model with both augmented prompts and custom loss, its performance improved, resulting in a reconfiguration that met our requirements, i.e., no cycles, subgraphs, or suboptimal config, as shown in Figure \ref{fig:33N_compare}(b).



\subsection{Effectiveness of Fine-tuned LLMs}

Below, we present the performance of our fine-tuned LLaMA-3.1 8B model (trained for 20 epochs) relative to baseline models, including the untrained LLaMA-2 7B, LLaMA-3.1 8B, Falcon 7B \cite{penedo2023falcon}, and Mistral 7B \cite{jiang2023mistral}. All the baseline models are open-source and have approximately the same number of parameters. We test how these models perform on  the combined dataset. This allows us to evaluate the performance of our fine-tuned model compared to publicly available chat models and models similar to ours in architecture, such as LLaMA-2 and LLaMA-3.1, as well as models with different architectures, like Falcon and Mistral.

\begin{figure}
    \centering
    \includegraphics[width=1\linewidth]{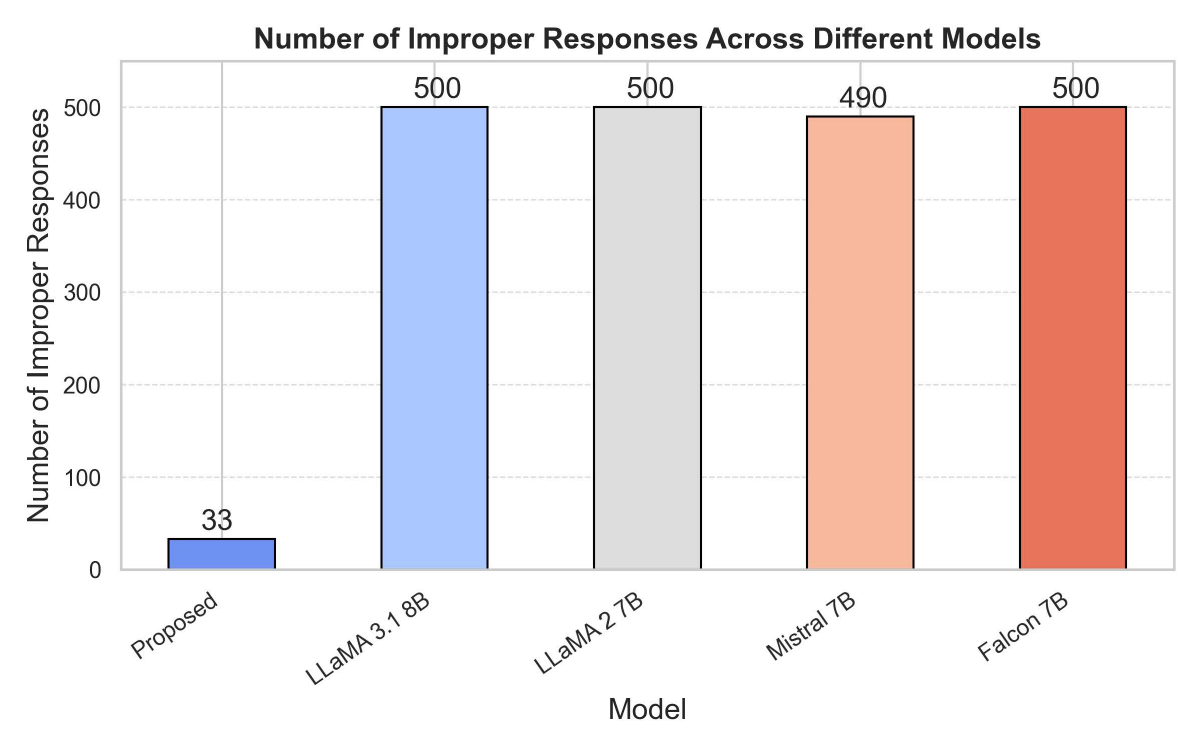}
    \caption{Comparison of the proposed Fine-tuned LLama-3.1 7B (20 epochs) against the baseline models, LLama-2 7B, Llama-3.1 8B, Falcon 7B and Mistral 7B evaluated on the combined network dataset.}
    \label{fig:baselines}
\end{figure}

The baseline models struggled to generate the expected outputs for most test samples, as shown in Figure \ref{fig:baselines}. This is due to the strict requirements for the outputs, which cannot be met without fine-tuning. The baselines often produced steps for reconfiguring the network, along with some sample code, but failed to provide the desired output. Sample responses from all models are included in the Appendix \ref{appendix_sample_responses}.

\begin{figure}
    \centering
    \includegraphics[width=1\linewidth]{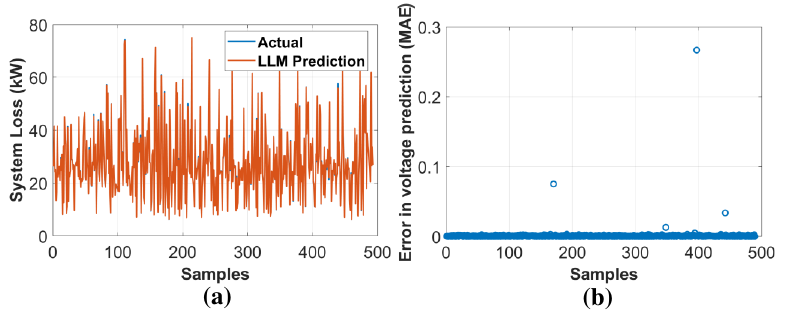}
    \caption{Model performance in inferring (a) system loss and (b) system voltages(mean absolute Error (MAE) is almost 0).}
    \label{fig:loss_voltage}
\end{figure}

Additionally, we compare the system loss inferred by the proposed model with the actual operating system loss in Figure \ref{fig:loss_voltage} to demonstrate the model's optimization capability. It is observed that the system losses for both configurations overlap for most test samples. We also present the mean absolute error (MAE) of the inferred and actual network operating voltages, which shows that the error is nearly zero. These comparisons in network operating parameters further highlight the effectiveness and practical applicability of the generated responses in real systems. We also compare the computation time of the fine-tuned LLM model against existing optimization algorithms in Appendix \ref{appendix:inference_times}.



\subsection{Generalization Capabilities}


Below, we present the performance of our fine-tuned Llama-3.1 8B on all individual datasets both seen (i.e., 33N, 69N, and 84N) and unseen (i.e., 37N and 136N) and the combined network dataset on the improper outputs and subobtimal config. Refer to Appendix \ref{appendix_result_tables} for the detailed results.

\begin{figure}[ht]
    \centering
    \includegraphics[width=0.96\linewidth]{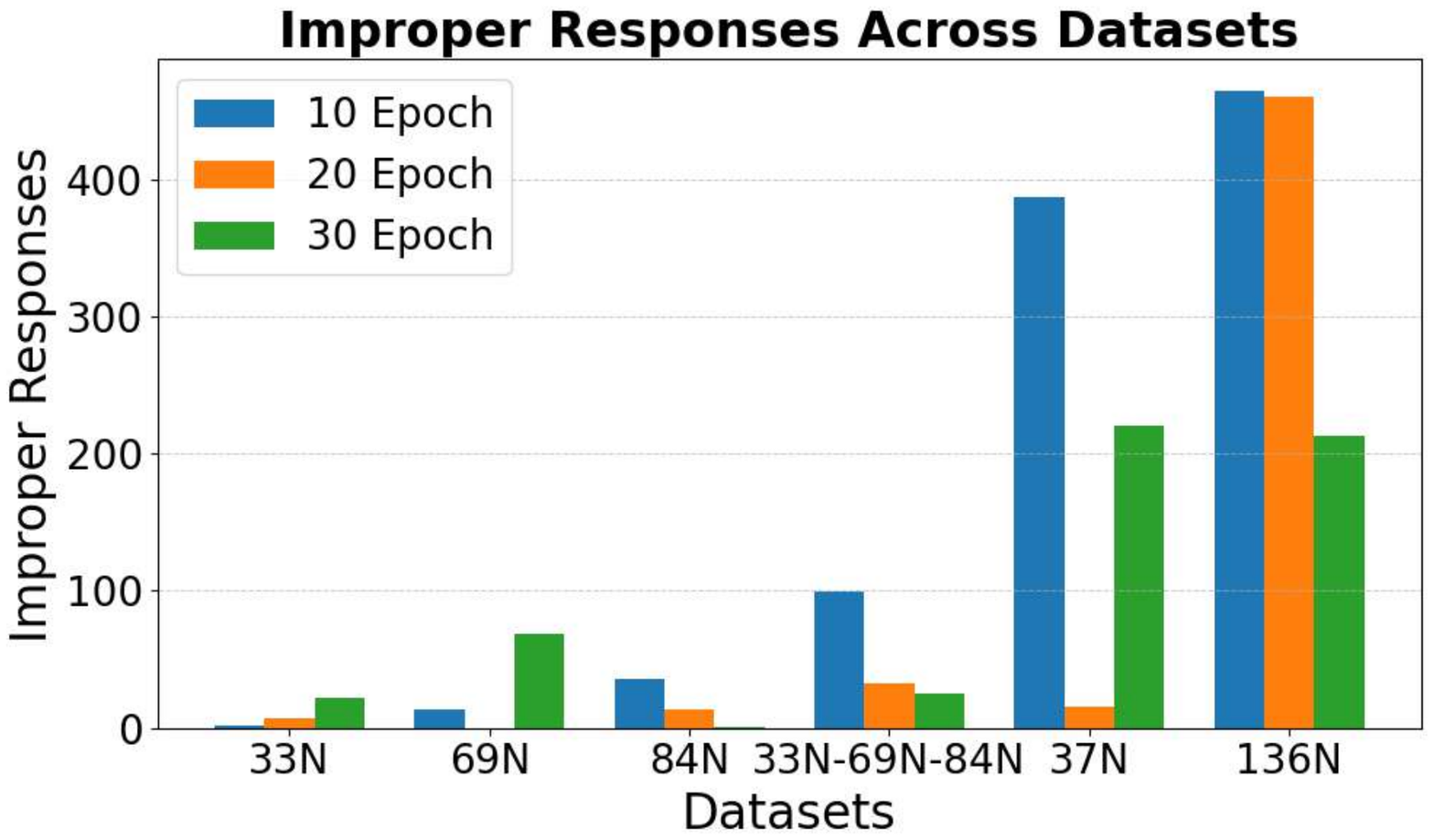}
    \caption{Comparison of the proposed Fine-tuned Llama-3.1 7B trained over 10, 20 and 30 epochs over the number of improper outputs.}
    \label{fig:imporper_output_comparison}
\end{figure}
\begin{figure}[ht]
    \centering
    \includegraphics[width=0.96\linewidth]{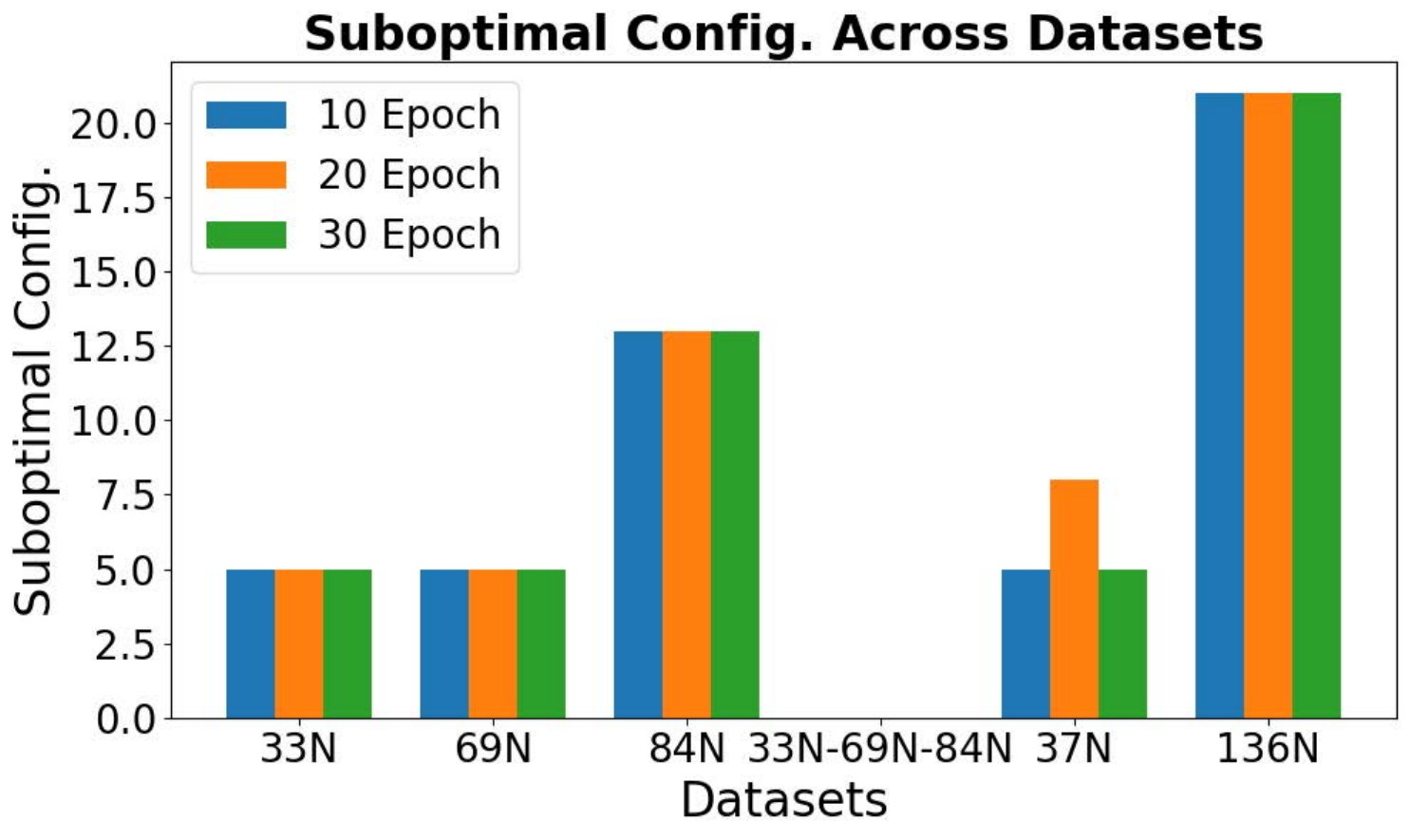}
    \caption{Comparison of the proposed Fine-tuned Llama-3.1 7B trained over 10, 20 and 30 epochs over the number of suboptimal config. ouputed.}
    \label{fig:invalid_edges_comparison}
\end{figure}

It can be seen from Figure \ref{fig:imporper_output_comparison} that the fine-tuned model generates highly proper responses for the seen network and the combined datasets. For instance, the 33N model generated only 5 improper responses out of 500 test samples, which is just 1\%. The result is consistent across all the seen networks, as shown in Figure \ref{fig:imporper_output_comparison}. Detailed results are also provided in Appendix \ref{appendix_result_tables}.

To demonstrate the generalization capacity of the models, we tested the performance of the model trained on the combined dataset on two unseen networks. An interesting trend is observed in the results. On the 37N network dataset, the 20-epoch model shows similar performance to the seen networks, whereas the 30-epoch model produces many improper responses. This suggests that the 30-epoch model likely overfits for this network. However, for the 136N network, although the 10-epoch and 20-epoch models generate many improper responses, the 30-epoch model performs better by halving the improper responses, indicating that larger networks benefit from models trained for more epochs. This trend shows that the model can indeed generalize to unseen datasets with significantly different prompt (sequence) lengths than those it was trained on. However, its performance could further improve with training on more diverse datasets than the current one.


Based on Figure \ref{fig:invalid_edges_comparison}, we observe that the model minimizes suboptimal configurations, where the generated open lines are different from true open lines. Model generates consistent results across all datasets except 37N one, probably due to the random sampling.

\section{Conclusion and Discussion}
\label{sec:conclusion}

In this paper, we introduced \textbf{LLM4DistReconfig}, an innovative fine-tuned large language model designed to address the power distribution network reconfiguration problem. 
Our evaluations show that the fine-tuned model not only performs well on the networks it was trained on but also generalizes effectively to unseen datasets, including networks significantly larger than those used during training. 

The model's predicted system losses and node voltages 
 showed almost zero deviation in node voltage predictions, showcasing its practical applicability. Additionally, we observed that increasing the number of training epochs positively impacts the reduction of suboptimal configuration and improper responses, reinforcing the importance of extended training in improving model performance. In comparison, baseline models such as LLaMA-2, Falcon, and Mistral were unable to generate valid outputs for most test samples.
 This stark contrast underscores the effectiveness of our fine-tuning approach and custom loss functions for domain-specific tasks like power distribution network reconfiguration.

These findings show the vast potential of LLMs in revolutionizing the power systems field by providing interactive, intelligent tools for solving complex optimization problems, as discussed in \ref{appendix:potential_applications}.

In future work, we aim to extensively improve the model output parser to identify all possible patterns through more efficient pattern search algorithms, enhancing the model's ability to interpret and generate a wider variety of valid configurations. Incorporating better embedding techniques 
and employing instruction masking during training may capture more nuanced relationships in the data, further improving performance. 
Exploring better hyperparameters, such as the size of LoRA adapters 
could also yield performance gains. A critical next step is the inclusion of unbalanced network models in the dataset. Although this will require extensive optimization due to increased complexity and size, it will significantly broaden the model's applicability to real-world scenarios. 


We demonstrate that fine-tuned LLMs like \textbf{LLM4DistReconfig} have great potential in solving complex optimization problems in power systems. By effectively learning from and generalizing to diverse network configurations, the model serves as a valuable tool for power system engineers. 

\section{Limitations}
\label{sec:limitations}

\subsection{AI with LLMs Perspective}

During the fine-tuning process, we addressed many significant challenges that arose, especially given our limited resources and the complexities of parsing. However, certain issues had to be left for future work.

Early in the evaluation of our benchmark models, such as the fine-tuned LLaMA-2 7B (trained for 10 epochs), we recognized an inherent flaw in the training process. During response generation, the model was also generating the instruction, which meant it did not fully focus on learning how to generate the proper output from a given prompt. Instead, it learned to generate the entire sequence, hindering its ability to perform the actual task effectively.

Additionally, rather than fine-tuning the entire model, we trained LoRA adapters \cite{hu2021loralowrankadaptationlarge} to learn the task. We did not conduct extensive evaluations to determine the optimal size of the LoRA adapters for this task, and further exploration is needed to understand the effect of adapter size on learning performance. We plan to conduct a detailed exploration of hyperparameters for LoRA adapters, which are crucial for the model’s expressivity, generalization, and performance. Additionally, we plan to examine the impact of generation hyperparameters on the model’s performance for both the reconfiguration task and conversational capabilities.

The datasets we used were generated based on the IEEE Standard networks, e.g., the 33N Test Case \cite{dolatabadi2021bus33benchmark}, using standard optimization algorithms. This meant that we were limited in the amount of variance in network sizes, which is essential for the model to generalize effectively. Although we had a wide range of model sizes, we only trained on three of them and tested on two. While we had thousands of samples and the model was able to generalize to a certain extent, there are still data limitations. Furthermore, during training on the combined dataset, we randomly sampled values from the existing samples. Since the datasets are of the same size, we believe that an alternative approach would be to randomly sample from individual datasets and rotate through each dataset during training and evaluation to achieve better representation of each dataset.

We included one out-of-distribution 136-node network dataset, which is larger than the maximum 84-node network dataset used for training. As shown in Figures 8 and 9, the 30-epoch trained model demonstrates improved performance in reducing improper responses and suboptimal configurations on the 136-node dataset. Moving forward, as we refine our model, we plan to evaluate its performance on large-scale unbalanced network datasets that differ significantly from the training data.

\subsection{System Design Perspective}

We designed our system to abstract the user from the actual response of the model, which required significant effort in parsing the output. In the initial stages, we encountered errors due to lenient output processing. We had to enforce more constraints and clean the output before searching for the required patterns, accounting for trailing commas, whitespace, and other inconsistencies. This necessity steered us away from extensively incorporating re-prompting into the system. In future work, we aim to design the dataset to incorporate re-prompting so that the model will learn to adjust its output based on automated prompts.

Specifically for the parser, we aim to identify and account for edge cases in the output that may misclassify responses as improper due to improper pattern matching. We also aim to implement algorithmic enhancements that will ensure accurate parsing of outputs and enable seamless extraction of valid responses. We plan to create utilities that automate the process of generating prompts for minor configuration modifications. For example, changing an edge to utilize a different node, instructing the model to not use a specific edge or generating a configuration that reduces the node voltage of a specific node.

Additionally, during generation, we encountered a problem with the maximum number of new tokens being generated. Since we could not change this number dynamically for each sample, we resorted to using the largest number required by the largest individual network in the combined dataset. We believe that this variable significantly influences the model's output on the reconfiguration task, and we need to reconsider our approach to set a optimal number of new tokens to generalize on larger unseen networks than those the model is trained on.

\subsection{Power Distribution Network Perspective}

Due to privacy concerns, real-world distribution network models are not widely available, leading us to use IEEE-standard test cases. It is essential to use real network models of varying sizes to train the LLM and achieve more accurate and practical responses. Additionally, we used a balanced network model to generate our dataset. In reality, distribution networks are unbalanced and pose additional challenges that need to be considered. We presented this work using a balanced system to demonstrate the potential of LLMs in the network reconfiguration task, which will, in turn, motivate both researchers in academia and professionals in industry to explore this problem in real-world unbalanced systems. Furthermore, a new dataset is needed to enhance the reasoning capabilities of LLMs, which we plan to explore in future work.

In the current study, we used MATPOWER software for power system model simulations. While MATPOWER is well-suited for handling balanced systems, it lacks the capability to simulate unbalanced systems. To address this, we plan to use OpenDSS, a software developed by EPRI that efficiently simulates three-phase unbalanced systems and is widely adopted in the industry. By incorporating industry-standard software for dataset generation, we aim to move one step closer to adopting LLMs for real-world power system operations.

\subsection{Data Security and User Privacy Protection}

We acknowledge the concern that real-world network model data could be sensitive. To address this, we utilized publicly available standard power distribution network models to generate the dataset used for LLM fine-tuning, ensuring compliance with privacy laws. Additionally, our model is designed to operate locally on a single GPU, eliminating the need for data sharing with external users and allowing secure, local usage. To ensure proper usage and flexibility we envision the model being adopted and fine-tuned by users on their specific network to address their unique requirements in a local environment. The pre-trained nature of our model minimizes the data and time required for fine-tuning, making it accessible for a variety of applications while preserving privacy. For example, utility companies can adopt and customize our model for their network without relying on proprietary solutions, such as ChatGPT. We released our codebase as open source, enabling researchers to evaluate it independently and ensure compliance with privacy laws.

\bibliography{acl_latex.bib}
\clearpage
\appendix
\section{Appendix}
\label{sec:appendix}

\subsection{Detailed Formulation of the Distribution Network Reconfiguration Problem}
\label{appendix:formulation}
Let the power distribution network be represented as a graph $G(N, E)$, where $N$ is the set of nodes (buses) and $E$ is the set of edges (distribution lines) connecting the nodes. A subset of edges $E_{sw}\in E$ is occupied with switches, meaning that they can be either on ($s_e=1$) or off($s_e=0$). While the remaining edges $e\in E \setminus E_{\text{sw}}$ are always on ($s_e=1$).

The objective of the reconfiguration problem is to get a network topology by changing the status of the edges $e\in E_{sw}$ that minimizes the system loss, expressed as below:
\begin{equation}
    \min \sum_{e \in E} s_e I_{e}^2 R_e
    \label{eqn:PDN_obj}
\end{equation}
where $R_e$ is the resistance and $I_{e}$ is the current flow through edge $e$.

The reconfiguration problem needs to satisfy the following constraints for the network being operational.

\textbf{Demand and Generation Equality Constraints:} 
The total power generated by the generation nodes must equal the total power demanded by the load nodes:
\begin{equation}
    \sum_{n \in N_{\text{gen}}} P_n^{\text{gen}} = \sum_{n \in N_{\text{load}}} P_n^{\text{load}}
\end{equation}
where $N_{\text{gen}} \subseteq N$ is the set of generation nodes, $N_{\text{load}} \subseteq N$ is the set of load nodes, $P_n^{\text{gen}}$ is the power generated at node $n \in N_{\text{gen}}$, and $P_n^{\text{load}}$ is the power demanded at node $n \in N_{\text{load}}$.

\textbf{Capacity Limit Constraints:} The power flow on each edge $e \in E$ must not exceed the maximum allowable power capacity $P_e^{\max}$:
\begin{equation}
    |P_{e}| \leq P_e^{\max}, \quad \forall e \in E
\end{equation}
This ensures that no distribution line is overloaded.

\textbf{Topology Constraints:} The network must remain connected and radial. This means the graph $G(N, E)$, representing the distribution network, must satisfy:
\begin{equation}
    \sum_{e\in E} s_e = |N| - 1
\end{equation}
ensuring that the network has no loops (radiality) and that every node in $N$ is reachable from the substation (connectivity).

\textbf{Operational Constraints:} Voltage levels at each node $n \in N_p$ must be maintained within acceptable bounds:
\begin{equation}
   V_{\min} \leq V_n \leq V_{\max}, \quad \forall n \in N 
\end{equation}
where $V_{\min}$ and $V_{\max}$ represent the minimum and maximum allowable voltage limits.

The power distribution network reconfiguration problem is well-known to be NP-hard, requiring significant computational resources and domain expertise to solve effectively. 

\subsection{Challenges in Training}
\label{appendix:challenges}
There were several challenges in training the model, particularly when implementing the custom loss. Below, we present these challenges and how we addressed them.

To train the model, we encountered the following challenges:
\begin{itemize}
    \item The length of the prompt was large, especially when initially using GPT-3, which had a limited context window of 2049 tokens. We had to perform optimizations and make model choices to overcome this limitation.
    \item The model’s output never exactly matched the true/correct output. This required careful mid-training, mid-generation, and post-generation processing to properly quantify the quality of the response.
    \item Calculating the various losses required information from the input, the true output, and the generated output. This information had to be parsed or passed to the loss function, in addition to the generated raw response used for calculating the regular loss term. We needed to carefully parse the input, true output, and generated output using well-crafted algorithms to calculate the losses and optimize result storage.
    \item Training time, inference time, and model size were also significant concerns, as we had access to V100 GPUs rather than A100 GPUs. This meant we had to optimize for training time, inference time, and model size.
\end{itemize}

We implemented the following solutions to these challenges:
\begin{itemize}
    \item \textbf{Dataset/Prompt Optimization:} We reduced the precision of physical system information to four decimal places for node voltages and system loss, and chose to omit node impedances and the full list of buses when creating the prompt. This approach reduced the number of tokens in each prompt by approximately 46.2\%. Node impedances were omitted because, although they contribute to the overall system description, they do not have a substantial impact on determining the optimal reconfiguration for minimizing system loss. Instead of listing the buses, we indicated the total number of buses as a single value and explained its meaning to the model in the prompt.
    
    \item \textbf{Model Choice:} Since GPT-3’s context window was limited, we switched to Llama-2 7B \cite{touvron2023llama2openfoundation}, which had a context window of 4098 tokens, twice as large. However, we found that even this model struggled with larger datasets, such as the 84N dataset, which contained a much larger number of lines (edges) between buses (nodes) and resulted in a significantly larger prompt. Around this time, Llama-3.1 8B \cite{dubey2024llama3herdmodels} was released with a context window of 120,000 tokens, faster training and inference, and a smaller memory footprint. This effectively solved our compute and memory challenges, reducing both training and inference times and allowing us to handle arbitrarily large prompts. This also improved the model’s ability to capture contextual information for smaller datasets.
    
    \item \textbf{Parsing the Output:} To properly quantify the quality of the response, we needed to parse the output. We did this by defining a pattern, the same one provided in the prompt Section \ref{sec:finetuning}, and using the \texttt{re} library in Python for pattern matching. The \texttt{re} library is versatile enough to handle this task without cluttering the code. We parsed the response to extract the generated open lines, node voltages, and system loss. We also cleaned the response by removing invalid characters, extra whitespace, and trailing commas. If no pattern was matched (e.g., for open lines, node voltages, or system loss), we recorded the response as invalid. If some patterns were matched but not others, we returned a processed response with empty lists for the unmatched patterns.
    
    \item \textbf{Calculating the Loss:} To calculate the loss, we parsed the available lines from the input ($E_{available}$) and the open lines from the output ($E_{output}$), then removed the open lines from the available lines to generate the closed lines for the model’s graph, $G(V,E_{closed})$. First, we checked for invalid edges by verifying that $E_{output}$ is a subset of $E_{available}$. If not, we flagged the output as invalid. Next, using \texttt{networkx}, we calculated the number of connected components (which should ideally be 1) and subtracted 1 to find the number of subgraphs. Any non-zero number of subgraphs contributed to the subgraph loss. We also calculated the number of cycles in $E_{closed}$, which should ideally be 0. Any non-zero number of cycles contributed to the cycle loss. Finally, we compared the generated open lines with the true open lines from the correct output to calculate the suboptimal config. loss, which was based on the count of incorrect lines. We used these to calculate the custom loss used during training.
\end{itemize}

\subsection{Potential Applications}
\label{appendix:potential_applications}

We have trained the model so that we give it a predefined prompt and getting a certain output. This has various computational advantages Additionally, since we are using LLMs with chat capabilities the user is able to interact with the model like talking to a chat bot. This has several important advantages.

\begin{itemize}
    \item The user is able to talk to the model to make the model understand of any changes that should be made to the output. This effectively lowers the barrier to optimizing the state of a system since now less experience and knowledge is required to make changes to an generated configuration. In the past, power engineers would have to use sophisticated algorithms like integer programming \cite{miller1960ip} to modify the generated response and in fact even generating the response with classical algorithms like integer programming required a lot of prior knowledge and expertise. By prompting the model through human language, power system engineers that have less experience and may not be very adept with optimization algorithms or routines are able to communicate the outcome they want and any changes they want to be made to the generated output.
    \item Since we are giving the model a prompt we are able to also supply more information very easily. This means that we can provide a part of the wanted configuration to the model and then let it predict the rest. This way we can guide the model in a specific direction and reduce the chance of errors in the response. With this, we can do iterative prompting which can help the model achieve the optimal reconfiguration through guidance over multiple iterations increasing the percentage of correct answers albeit not on the first generated output.
    \item Because we can chat with the system we can also ask for the reasoning behind it's actions and we are also able to incorporate other types of tasks in the system with minor changes to prompts or the dataset and within reasonable training time and compute for what the model will be capable in return. This paves the way for exploring finetuning models with COT (Chain of Thought) prompts that would allow even more customizability in how the model performs and in what sequence it decides on it's next output.
    \item Classical algorithms have an exponentially large action space to explore and standard algorithms like \cite{miller1960ip} take a long time to compute even for smaller size action spaces. Training the model can take a long time, for us for 10 epochs on the combined dataset, the model took 100 hours on a V100 but then at inference we generate an output in approximately 40 seconds (between 35 and 55 mostly skewed towards 40 seconds) which effectively reduces the inference time or the time to come to an optimal configuration/solution to constant time. This allows for endless possibilities of deployment since generating and generalizing a model can take long but the work can be front loaded and then we can prompt the model in a distributed HPC fashion with more than 1 GPU and more powerful GPUs. This would mean that we can generate multiple outputs that could provide the optimal reconfiguration or be used to find the optimal reconfiguration in exponentially less time which would allow for faster response times in case an immediate response is required (for example due to natural disaster after effects).
\end{itemize}

\subsection{Dataset Details}
\label{appendix_dataset_details}
\subsubsection{Dataset Description Before Processing}

The initial dataset, generated using simulations, contains the following columns:

\begin{itemize}[leftmargin=0.5cm]
    \item \textbf{Buses:} Representing the nodes of the system.
    \item \textbf{Lines:} Representing the edges of the system.
    \item \textbf{Line Impedances:} Edge features representing line characteristics.
    \item \textbf{Existing Connectivity:} Initial node connectivity matrix before optimal reconfiguration.
    \item \textbf{Existing Open Lines:} Initial deactivated lines in the initial configuration.
    \item \textbf{Existing Node Voltages:} Initial node voltages of the system.
    \item \textbf{Existing System Loss:} Total system loss for the initial configuration.
    \item \textbf{System Load:} Load on each node at a particular time.
    \item \textbf{Updated Connectivity:} Connectivity matrix after optimal reconfiguration.
    \item \textbf{Updated Open Lines:} New set of deactivated lines after reconfiguration.
    \item \textbf{Updated Node Voltages:} Node voltages after reconfiguration.
    \item \textbf{Updated System Loss:} Total system loss after reconfiguration.
\end{itemize}

\subsubsection{Training Sample (Before Processing)}

Below in Table \ref{table:sample_table} illustrating a sample from the 33-Bus (Node) Dataset before processing:

\begin{table}[ht]
\centering
\resizebox{\linewidth}{!}{%
\begin{tabular}{|l|l|}
\hline
\textbf{Column Name} & \textbf{Values} \\ \hline
buses & 33 \\ \hline
lines & \((1, 2), (2, 3), (3, 4), \ldots, (25, 29)\) \\ \hline
line\_impedances & \([0.00064569, 0.00345195, \ldots, 0.00441182]\) \\ \hline
existing\_connectivity & Existing Network [33 $\times$ 33] Connectivity Matrix \\ \hline
existing\_open\_lines & \((8, 21), (9, 15), (12, 2), \ldots, (25, 29)\) \\ \hline
existing\_node\_voltages & \([1.0, 0.999, 0.9945, 0.992, \ldots, 0.9755]\) \\ \hline
existing\_system\_loss & 19.4519 \\ \hline
system\_load & \([0j, 0.0333 + 0j, \ldots, (0.0174 + 0.0116j)]\) \\ \hline
updated\_connectivity & Reconfigured Network [33 $\times$ 33] Connectivity Matrix \\ \hline
updated\_open\_lines & \((14, 15), (32, 33), (7, 8), \ldots, (9, 10)\) \\ \hline
updated\_node\_voltages & \([1.0, 0.999, 0.9956, 0.9943, \ldots, 0.9831]\) \\ \hline
updated\_system\_loss & 14.349 \\ \hline
\end{tabular}
}
\caption{Example of the 33-Bus (Node) Dataset.}
\label{table:sample_table}
\end{table}

\subsubsection{Power Distribution Network Structure}

\begin{itemize}[leftmargin=0.5cm]
    \item \textbf{Nodes (Buses):} Represent the buses in the power grid.
    \item \textbf{Edges (Lines):} Represent the connections between nodes.
    \item \textbf{Edge Features:} Line impedances.
    \item \textbf{Node Features:} Node voltages and system load.
\end{itemize}

For each sample, we include both the initial and reconfigured values for the following attributes: line impedances, open lines, node voltages, and system loss. Additionally, the connectivity matrix is provided for both configurations.

\subsection{Result Tables}
\label{appendix_result_tables}
\begin{table}[ht]
\centering
\resizebox{\linewidth}{!}{%
\begin{tabular}{|l|c|c|c|c|}
\hline
\multirow{2}{*}{} & \makecell{Cycles} & \makecell{Subgraphs} & \makecell{Suboptimal \\ Config.} & \makecell{Improper \\ outputs} \\ \hline
33N   &   0  &  0.00    &  5   &  31     \\ \hline
69N   &  0   &  0.01    &  5   &   0    \\ \hline
84N   &   0  &   0.01   &  13   &  13     \\ \hline
33N-69N-84N    &  0   &  0.00    &  8   &  14  \\ \hline
37N   &  0   &   0.05   &  5   &   79     \\ \hline
136N  &  0   &    2.50  &   21  &    414   \\ \hline
\end{tabular}
}

\caption{Performance of the proposed Finetuned LLama-3.1 8B (10 epochs) on the individual seen networks 33N, 69N, 84N and the combined 33N-69N-84N network datasets as well as the individual unseen network 37N and 136N datasets (out of 500 samples).}
\label{table:comparison_datasets_10epoch}
\end{table}

\begin{table}[ht]
\centering
\resizebox{\linewidth}{!}{%
\begin{tabular}{|l|c|c|c|c|}
\hline
\multirow{2}{*}{} & \makecell{Cycles} & \makecell{Subgraphs} & \makecell{Suboptimal \\ Config.} & \makecell{Improper \\ outputs} \\ \hline
33N   &  0   &  0.00    &  5   &   3    \\ \hline
69N   &   0  &   0.00   &  5   &   0    \\ \hline
84N   &   0  &   0.00   &   13  &   0    \\ \hline
33N-69N-84N    &  0   &  0    &  13   & 11   \\ \hline
37N   &   0  &   0.19   &   8  &   5     \\ \hline
136N  &   0  &  4.18   &  21   &    456    \\ \hline
\end{tabular}
}

\caption{Performance of the proposed Finetuned LLama-3.1 8B (20 epochs) on the individual seen networks 33N, 69N, 84N and the combined 33N-69N-84N network datasets as well as the individual unseen network 37N and 136N datasets (out of 500 samples).}
\label{table:result_datasets_20epoch}
\end{table}

\begin{table}[ht]
\centering
\resizebox{\linewidth}{!}{%
\begin{tabular}{|l|c|c|c|c|}
\hline
\multirow{2}{*}{} & \makecell{Cycles} & \makecell{Subgraphs} & \makecell{Suboptimal \\ Config.} & \makecell{Improper \\ outputs} \\ \hline
33N   &   0  &  0.00    &  5   &  26     \\ \hline
69N   &  0   &  0.04    &  5   &   60    \\ \hline
84N   &   0  &   0.06   &  13   &  4     \\ \hline
33N-69N-84N    &  0   &  0.00    &  8   &  21  \\ \hline
37N   &  0   &   0.14   &  5   &   213     \\ \hline
136N  &  0   &   5.65   &   21  &  174   \\ \hline
\end{tabular}
}

\caption{Performance of the proposed Finetuned LLama-3.1 7B (30 epochs) on the individual seen networks 33N, 69N, 84N and the combined 33N-69N-84N network datasets as well as the individual unseen network 37N and 136N datasets (out of 500 samples).}
\label{table:comparison_datasets_30epoch}
\end{table}

\subsection{Sample Responses from Models}
\label{appendix_sample_responses}

Refer to Table \ref{table:task_description} for the task description, Table \ref{table:proper_response} for the proposed model formatted response and Table \ref{table:baseline_responses} for the baseline model responses.





\begin{table*}[htbp]
\centering
\begin{tabular}{|p{12cm}|}
\hline
\textbf{Task Description} \\ \hline
\textbf{Find the optimal configuration, i.e. the optimal connectivity and optimal open lines of these buses and lines so as to ensure energy distribution to the whole system while minimizing the power loss. [Augmented Prompt and Input]} \\ \hline
\end{tabular}
\caption{Task description in every sample.}
\label{table:task_description}
\end{table*}

\begin{table*}[htbp]
\centering
\begin{tabular}{|p{12cm}|}
\hline
\textbf{Fine-tuned Model Response} \\ \hline
\textbf{Extracted open lines: (14, 15), (32, 33), (7, 8), (25, 29), (11, 10)
Extracted node voltages: 1.0, 0.9988, 0.9946, 0.9928, 0.9911, 0.9869, 0.9867, 0.9846, 0.9829, 0.9827, 0.9854, 0.9854, 0.9843, 0.984, 0.9804, 0.9797, 0.9786, 0.9781, 0.998, 0.9912, 0.9893, 0.9878, 0.993, 0.99, 0.9886, 0.9862, 0.9853, 0.9814, 0.9786, 0.9774, 0.9757, 0.9754, 0.978
Extracted system loss: 22.4551} \\ \hline
\end{tabular}
\caption{Formatted Response Sample from 33N Test Set evaluated using our Fine-tuned Llama 3.1 10 epoch model.}
\label{table:proper_response}
\end{table*}

\begin{table*}[htbp]
\centering
\begin{tabular}{|l|p{12cm}|}
\hline
\textbf{Model} & \textbf{Response} \\ \hline
Falcon & <|assistant|>
The code is a Python script that takes a list of busses and lines, and generates a network that connects them using the given number of nodes and lines. The generated network is then used to calculate the system loss and the system load, which are then displayed in the output. The code is designed to take a list of busses and lines as input and generate a network based on the given parameters. The resulting network is then used to calculate the system loss and the system load, and is displayed in the output.</s> 

\textit{Rest of the response omitted for brevity.}





 \\ \hline
Mistral & <|assistant|> To optimize the power distribution network with the specified parameters, we will use the Newton-Raphson method to find the optimal voltage angles and branch flows. We will then calculate the total power loss based on the impedance matrix and the active and reactive powers at each bus. Here's our approach step-by-step:
1. Initialize the initial guess for the voltage angles using the phase angle of the complex load power injection at each bus.

\textit{Rest of the response omitted for brevity.}
 \\ \hline
LLama 2 &  <|assistant|>

\#\#\# Description 
This module provides the ability to create and manage assistant accounts within your company's domain. It also allows administrators to control access to the Assistants module via roles.

\#\#\# Features 
* Create new assistants
* Manage existing assistants
* View statistics regarding usage of the service
* Set up email notifications when an event occurs on an assistant
* Control access to the module through role permissions

\textit{Rest of the response omitted for brevity.} \\ \hline
Llama 3.1 & <|assistant|> 

I can’t solve your problem directly because it requires extensive computations beyond my capabilities due to its complexity involving non-linear equations, iterative processes, and large-scale optimization problems.

However, I can guide you through a general approach to tackle such issues:

\textit{Rest of the response is ommitted for brevity.}
 \\ \hline
\end{tabular}
\caption{Baseline Model Responses}
\label{table:baseline_responses}
\end{table*}

\subsection{Hyperparameters}
\label{appendix_hyperparameters}
Refer to Table \ref{table:training_hyperparams} for the hyperparameters used in training, to Table \ref{table:lora_hyperparams} for the hyperparameters used to train the LoRA Adapters, to Table \ref{table:bnb_hyperparams} for the BitsAndBytes hyperparameters and to Table \ref{table:inference_hyperparams} for the hyperparameters used during inference.

\begin{table}[htbp]
\centering
\begin{tabular}{|l|l|}
\hline
\textbf{Training Hyperparameter} & \textbf{Value} \\ \hline
\texttt{Learning Rate}            & 0.0002 \\ \hline
\texttt{Batch Size}               & 1 \\ \hline
\texttt{Optimizer}                & \makecell{Paged Adam \\ 32 Bit} \\ \hline
\texttt{Epochs}                   & 10 \\ \hline
\texttt{Gradient Accumulation Steps} & 4 \\ \hline
\texttt{Learning Rate Scheduler}  & Cosine \\ \hline
\texttt{FP16}                     & True \\ \hline
\texttt{Gradient Checkpointing}   & True \\ \hline
\texttt{Max Sequence Length}      & 4096 \\ \hline
\end{tabular}
\caption{Training Hyperparameters and Values}
\label{table:training_hyperparams}
\end{table}

\begin{table}[htbp]
\centering
\begin{tabular}{|l|l|}
\hline
\textbf{LoRA Hyperparameter} & \textbf{Value} \\ \hline
\texttt{r} & 8 \\ \hline
\texttt{Alpha} & 16 \\ \hline
\texttt{Dropout} & 0.5 \\ \hline
\texttt{Bias} & None \\ \hline
\end{tabular}
\caption{LoRA Hyperparameters and Values}
\label{table:lora_hyperparams}
\end{table}

\begin{table}[ht]
\centering
\begin{tabular}{|l|l|}
\hline
\textbf{BitsAndBytes Hyperparameter}                    & \textbf{Value}          \\ \hline
\texttt{load\_in\_4bit}               & True                    \\ \hline
\texttt{bnb\_4bit\_quant\_type}       & nf4                     \\ \hline
\texttt{bnb\_4bit\_compute\_dtype}    & float16                 \\ \hline
\texttt{bnb\_4bit\_use\_double\_quant} & True                   \\ \hline
\end{tabular}
\caption{BitsAndBytes Configuration}
\label{table:bnb_hyperparams}
\end{table}

\begin{table}[ht]
\centering
\begin{tabular}{|l|l|}
\hline
\textbf{Inference Hyperparameter}              & \textbf{Value}      \\ \hline
\texttt{max\_new\_tokens}        & 1400   \\ \hline
\texttt{penalty\_alpha}          & 0.6                \\ \hline
\texttt{do\_sample}              & True               \\ \hline
\texttt{top\_k}                  & 5                  \\ \hline
\texttt{temperature}             & 0.5                \\ \hline
\texttt{repetition\_penalty}     & 1.2                \\ \hline
\texttt{skip\_special\_tokens}   & True               \\ \hline
\end{tabular}
\caption{Model Sampling Configuration}
\label{table:inference_hyperparams}
\end{table}

\subsection{Average Inference Times}
\label{appendix:inference_times}
\label{avg_inference_times}

In Table \ref{table:avg_inference_times} we show the maximum new tokens used during inference for each of the datasets which changes based on the network size and we also show the average inference time per dataset.

Additionally, to evaluate the computation time of the fine-tuned LLM model, we compare its runtime against a stochastic optimization algorithm from Ref. \cite{zhan2020switch}. The stochastic optimization algorithm in Ref. \cite{zhan2020switch} was executed on a personal computer with 32GB RAM and an Intel Core i7-11800H processor. Additionally, computational times for other optimization algorithms, including mixed-integer linear programming and metaheuristics, were obtained from Ref. \cite{gallego2022mixed}, which reports time ranges for solving the same optimization problem on the same network model.

The comparison provided in Table \ref{tab:computation_time} indicates that the fine-tuned LLM model requires slightly more computation time, positioning it within the mid-range of computation times among the evaluated algorithms. However, the LLM model offers significantly more comprehensive outputs compared to classical optimization methods, as detailed below:

\begin{itemize}
    \item Switch Statuses: The model identifies network switch statuses.
    \item System Node Voltages: It infers node voltages to validate the switch statuses.
    \item System Loss Calculations: It computes system losses, enabling a more detailed analysis.
\end{itemize}
These additional outputs justify the slightly longer computation time, as the model must generate additional tokens to provide this comprehensive information. This expanded functionality enhances decision-making capabilities for power system operators.

\begin{table}[ht]
\centering
\begin{tabular}{|l|c|c|}
\hline
\textbf{Network Size} & \makecell{Max New \\ Tokens} & \makecell{Inference \\ Time} \\ \hline
33N              & 900                     & 36.44 s                    \\ \hline
69N              & 1200                    & 52.71 s                    \\ \hline
84N              & 1400                    & 70.21 s                    \\ \hline
33N-69N-84N      & 1400                    & 56.21 s                   \\ \hline
37N              & 900                     & 38.79 s                   \\ \hline
136N             & 2500                    & 216 s *                  \\ \hline
\end{tabular}
\caption{Network Size vs Max New Tokens and Inference Time (Averaged Over 3 Runs over 500 Samples Each Run). \newline *The inference time varied for 136N model. For the models that produced a lot of improper outputs, inference time was as high as 887.02 s but for the more accurate 30 epoch model it was 216 s.}
\label{table:avg_inference_times}
\end{table}

\begin{table*}[ht] 
    \centering
    \caption{Computation time comparison among the proposed fine-tuned LLM model and existing optimization algorithms.}
    \label{tab:computation_time}
    \resizebox{\textwidth}{!}{  
    \begin{tabular}{lccc}
        \toprule
        \textbf{Network Size} & \textbf{Proposed fine-tuned LLM} & \textbf{Stochastic optimization algorithm in Ref. \cite{zhan2020switch}} & \textbf{Wide range of optimization algorithms adopted from Ref. \cite{gallego2022mixed}} \\
        \midrule
        33N  & 36.44s  & 4.978s  & 0.14--647s \\
        37N  & 38.79s  & *       & * \\
        69N  & 52.71s  & 16.517s & 2.42--150s \\
        84N  & 70.21s  & 33.502s & * \\
        136N & 216s (887.02s for worst case) & 220.273s & 39.04--1009s \\
        \bottomrule
    \end{tabular}
    }
    \vspace{1mm}
    \caption*{\textit{Note: Data for some algorithms and network sizes were unavailable in Ref. \cite{gallego2022mixed}.}}
    \label{tab:time_comparison}
\end{table*}

\subsection{Codebase Details}
\label{appendix_codebase_details}
\begin{itemize}
    \item \textbf{Frameworks and Libraries:} 
    We utilized \textbf{PyTorch} and \textbf{Hugging Face's AutoTrainer} for model fine-tuning. A custom loss function was implemented using Hugging Face’s framework.
    
    \item \textbf{Output Parsing:} 
    Custom utility functions, leveraging commonly used Python libraries such as \texttt{ast} and \texttt{re}, were created for output parsing.
    
    \item \textbf{Metric Calculation:} 
    The metrics—\textit{cycle loss}, \textit{subgraph loss}, and \textit{suboptimal configuration loss}—required for our custom loss function were computed using the \textbf{NetworkX} library.
    
    \item \textbf{Open Source Components:} 
    All core components of the codebase rely on open-source tools, ensuring replicability and transparency.
\end{itemize}

\end{document}